\documentclass[lettersize,journal]{IEEEtran}
\usepackage{amsmath,amsfonts}
\usepackage{algorithmic}
\usepackage{algorithm}
\usepackage{array}
\usepackage[caption=false,font=normalsize,labelfont=sf,textfont=sf]{subfig}
\usepackage{textcomp}
\usepackage{stfloats}
\usepackage{url}
\usepackage{verbatim}
\usepackage{graphicx}
\usepackage{cite}
\usepackage{tabularx} %for helix table
\usepackage{booktabs}
\usepackage[dvipsnames]{xcolor}

\begin{document}

\title{Py-DiSMech: A Scalable and Efficient Framework for Discrete Differential Geometry-Based Modeling and Control of Soft Robots}

\author{Radha Lahoti$^{1, *}$, Ryan Chaiyakul$^{2, *}$ and M. Khalid Jawed$^{1}$

        % <-this % stops a space
\thanks{This research was funded by the National Science Foundation (award numbers: CMMI-2209782, CAREER-2047663, CMMI-2332555) and the National Institutes of Health (1R01NS141171-01).}% <-this % stops a space
\thanks{$^{1}$ Department of Mechanical and Aerospace Engineering, University of California, Los Angeles (UCLA), CA 90095, USA.
        {\tt\small radhalahoti@ucla.edu, khalidjm@seas.ucla.edu}}%  
\thanks{$^{2}$ Department of Electrical and Computer Engineering, University of California, Los Angeles (UCLA), CA 90095, USA.
        {\tt\small ryanchaiyakul@ucla.edu}}%  
\thanks{$^{*}$ The authors contributed equally.}
}

\markboth{Journal of \LaTeX\ Class Files,~Vol.~14, No.~8, August~2021}%
{Shell \MakeLowercase{\textit{et al.}}: A Sample Article Using IEEEtran.cls for IEEE Journals}

\IEEEpubid{0000--0000/00\$00.00~\copyright~2021 IEEE}

\maketitle

\begin{abstract}
High-fidelity simulation has become essential to the design and control of soft robots, where large geometric deformations and complex contact interactions challenge conventional modeling tools. Recent advances in the field demand simulation frameworks that combine physical accuracy, computational scalability, and seamless integration with modern control and optimization pipelines. In this work, we present Py-DiSMech, a Python-based, open-source simulation framework for modeling and control of soft robotic structures grounded in the principles of Discrete Differential Geometry (DDG). By discretizing geometric quantities such as curvature and strain directly on meshes, Py-DiSMech captures the nonlinear deformation of rods, shells, and hybrid structures with high fidelity and reduced computational cost. The framework introduces (i) a fully vectorized NumPy implementation achieving order-of-magnitude speed-ups over existing geometry-based simulators; (ii) a penalty-energy-based fully implicit contact model that supports rod–rod, rod–shell, and shell–shell interactions; (iii) a natural-strain-based feedback-control module featuring a proportional–integral (PI) controller for shape regulation and trajectory tracking; and (iv) a modular, object-oriented software design enabling user-defined elastic energies, actuation schemes, and integration with machine-learning libraries.
% such as PyTorch and TensorFlow.
Benchmark comparisons demonstrate that Py-DiSMech substantially outperforms the state-of-the-art simulator Elastica in computational efficiency while maintaining physical accuracy. Together, these features establish Py-DiSMech as a scalable, extensible platform for simulation-driven design, control validation, and sim-to-real research in soft robotics. 
\end{abstract}

\begin{IEEEkeywords}
soft robotics, discrete differential geometry, rods, shells, contact modeling, feedback control
\end{IEEEkeywords}

\section{Introduction}
\IEEEPARstart{A}{ccurate} and efficient simulation tools have become indispensable in modern robotics research. They serve as digital testbeds that allow researchers to prototype, validate, and iterate control strategies and designs before engaging in resource-intensive physical experimentation. As robotic systems grow in complexity, especially in domains such as soft robotics, the role of high-fidelity simulation in accelerating scientific progress has become increasingly central. A reliable simulator not only reduces experimental costs but also shortens development cycles and enables scalable exploration of design and control paradigms.

Simulating soft robotic systems, however, introduces unique challenges. Unlike rigid-body systems, soft robots undergo large deformations governed by continuum mechanics. Capturing these deformations accurately while maintaining computational efficiency requires modeling frameworks that can handle geometric and material nonlinearities in a stable and tractable manner. Despite their importance, general-purpose robotic simulators often fall short in this regard. Many rely on simplified lumped-mass approximations or rigid-link networks, which fail to capture the nuanced behavior of soft, compliant structures. More sophisticated tools, such as those based on Finite Element Analysis (FEA), achieve improved accuracy but are often computationally prohibitive and difficult to adapt to robotics-specific use cases.

Adding to this complexity is modeling soft contact interactions between deformable bodies such as rods, shells, and hybrid structures. Robust contact handling under large deformations remains a critical barrier to achieving physically plausible simulations of tasks such as gripping, impact, or locomotion through constrained environments. Contact models must be not only accurate but also stable and compatible with large time steps to be viable in dynamic simulations.

To address these challenges, Discrete Differential Geometry (DDG) has emerged as a compelling modeling framework. DDG methods discretize geometric quantities (such as curvature and strain) directly on meshes, enabling the simulation of deformable bodies with fewer degrees of freedom and improved computational efficiency. Moreover, DDG is particularly well-suited for implementing penalty energy-based implicit contact models—yielding physically accurate and numerically stable behavior even in contact-rich scenarios. Prior work~\cite{choi2023dismech, lahoti2025matdismech} has demonstrated the promise of DDG-based methods for modeling soft rods and shells with high fidelity.

\IEEEpubidadjcol % to prevent overlap with \IEEEpubid
Building on this foundation, we introduce Py-DiSMech, a Python-based, open-source simulation framework for soft robots grounded in the principles of DDG. In recent years, Python has become the de facto language for scientific computing, machine learning, and AI research—supported by ecosystems such as PyTorch, TensorFlow, and JAX. A Python-native simulator thus broadens accessibility for the robotics community while enabling seamless integration with modern ML pipelines, including zero-copy interoperability between NumPy arrays and deep learning frameworks. Py-DiSMech extends prior geometry-based soft-robot simulation frameworks in several key ways:

% \IEEEpubidadjcol % to prevent overlap with \IEEEpubid

\begin{enumerate}
    \item \textbf{Performance through Vectorization:} Py-DiSMech leverages NumPy-based vectorized computations in place of loop-based stencil evaluations, leading to a significant speed-up in simulation performance as compared to existing state-of-the-art methods. This enables larger-scale simulations and faster iteration times.
    \item \textbf{Implicit Contact implementation for hybrid Rod-Shell Structures:} Py-DiSMech generalizes the penalty energy-based fully implicit contact model~\cite{choi_imc_2021} to support not just rod–rod contact (as in Elastica~\cite{gazzola2018}, DisMech~\cite{choi2023dismech} and MAT-DiSMech~\cite{lahoti2025matdismech}), but also shell–shell and rod–shell interactions. This broadens the scope of applications and allows for realistic simulations of hybrid soft structures in contact-rich tasks.
    \item \textbf{Feedback Control Framework:} Py-DiSMech integrates natural-strain-based feedback control directly into the simulation pipeline, enabling closed-loop shape regulation and trajectory tracking. A proportional–integral (PI) controller is provided as a built-in example, and the modular design allows straightforward extension to advanced methods such as model-predictive or optimal control—supporting rapid prototyping and sim-to-real transfer studies.
    \item \textbf{Object-Oriented and Extensible Software Design:} Py-DiSMech is structured in a modular, object-oriented fashion, enabling researchers to customize and extend the framework easily. As a case study, we demonstrate how users can incorporate a custom elastic energy term—showcasing the framework’s flexibility in modeling beyond the built-in rod and shell dynamical models.
    \item \textbf {Seamless ML Integration and Open-Source Accessibility:} Py-DiSMech’s native Python implementation enables easy integration with machine learning libraries like PyTorch and TensorFlow, facilitating hybrid modeling approaches that combine physics with data-driven components. Moreover, being fully open-source and free from proprietary dependencies, it offers broader accessibility compared to MATLAB-based tools or licensed commercial software.
\end{enumerate}

\begin{figure*}[t!]
\centerline{\includegraphics[clip, trim=0cm 1cm 0cm 0cm, width =\textwidth]{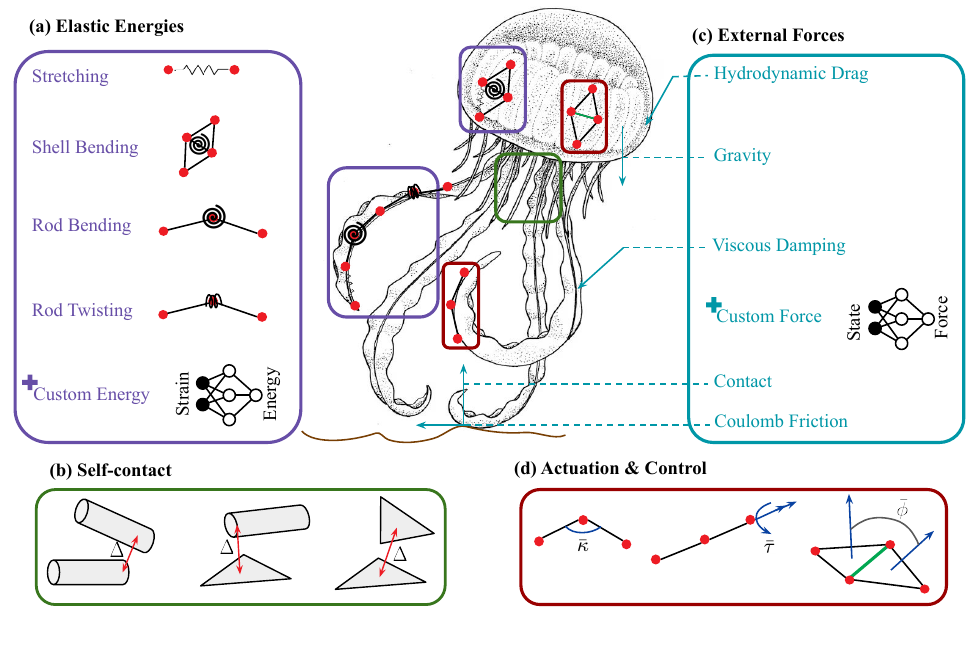}}
\caption{Key functional components of Py-DiSMech: (a) elastic energy formulations, (b) self-contact modeling, (c) external force integration, and (d) actuation and control through strain- or curvature-based inputs}
\label{fig::first}
\end{figure*}

Together, these contributions establish Py-DiSMech as a comprehensive and extensible platform for simulating the dynamics of soft robotic systems. By combining the geometric rigor of discrete differential geometry with the computational efficiency of Python-based vectorization, Py-DiSMech bridges the gap between physical fidelity and performance, enabling simulation-driven design and optimization of complex soft robots. Moreover, its integrated control framework extends the simulator’s utility beyond passive modeling, providing a foundation for the design, testing, and validation of feedback and learning-based control strategies. Figure~\ref{fig::first} summarizes the key capabilities of Py-DiSMech.

The remainder of this paper is organized as follows. Section~\ref{sec: related_works} reviews related work and motivates the relevance of our approach in the context of current research. Section~\ref{sec: dynamics} presents the dynamical modeling techniques employed in our simulator. Section~\ref{sec: software} describes the software framework and its usage. In Section~\ref{sec: comparison}, we compare Py-DiSMech with the state-of-the-art soft robotics simulator Elastica, focusing on computational speed. Section~\ref{sec: results} showcases representative simulations that highlight the framework’s capabilities and versatility. Section~\ref{sec: control} details a closed-loop PI control strategy for regulation and tracking, and Section~\ref{sec: conclusions} concludes the manuscript.

\section{Related Work}
\label{sec: related_works}

Simulation tools for soft robots can broadly be categorized into three classes:
(1) Lumped-mass-based models, which approximate deformable bodies using point masses connected by springs;
(2) Finite Element Analysis (FEA)-based models, grounded in continuum mechanics and partial differential equations; and
(3) Geometry-based models, which simplify dynamics by leveraging the structural slenderness through reduced-order formulations.

These classes represent a continuum of trade-offs between computational efficiency and physical accuracy. Lumped-mass models offer real-time performance but at the expense of realism. FEA-based methods capture rich physical detail but are often prohibitively slow for dynamic or closed-loop tasks. Geometry-based approaches aim to balance these extremes by modeling dominant deformation behaviors—especially in slender rod or shell-like structures—at reduced computational cost. In the following, we review representative tools across these categories, highlighting their modeling assumptions, performance characteristics, and limitations in the context of soft robot simulation.

Lumped-mass-based simulators such as MuJoCo~\cite{todorov2012mujoco} and PyBullet~\cite{coumans2021} achieve speed by approximating soft structures with mass-spring networks or compliant rigid-link chains. SoMo~\cite{graule2020somo} is a framework that builds on PyBullet to support continuum manipulators, enabling real-time simulation for soft arm control and learning tasks. However, these models lack the fidelity needed to capture complex material behavior or large-deformation interactions, such as self-contact and wrinkling, limiting their utility in physically grounded soft robot design. 

At the other end of the spectrum, FEA-based simulators model soft bodies using high-order continuum mechanics. SOFA~\cite{faure2012} supports a range of element types, including Cosserat rods via plugins, and provides tools for simulating actuation mechanisms like cable pulls and pressure chambers. Project CHRONO~\cite{Chrono2016} is another FEA-based simulator which has been shown to handle frictional contact in granular media effectively~\cite{chrono_contact2017}. Isaac Sim~\cite{isaac-sym}, developed by NVIDIA, extends PhysX~\cite{nvidia2024_physx_release} with FEM-based soft body dynamics and improves usability in robotics contexts. However, its support for actuation is limited—soft bodies are passive.

Commercial software like ANSYS~\cite{ansys} and ABAQUS~\cite{abaqus} offer highly sophisticated modeling capabilities but are typically used for offline analysis due to their computational intensity. ABAQUS, in particular, is regarded as a state-of-the-art tool for high-fidelity, FEA-based simulations, offering exceptional customizability through user-defined subroutines for materials, elements, forces, and boundary conditions. Notably, while DDG-based approaches offer computationally efficient alternatives for soft robot simulation, no existing DDG-based software provides a comparable level of customization or expandability. Motivated by this, we aim to develop a DDG-based soft robot simulator that fills this gap, offering the same degree of extensibility as ABAQUS does for FEA, but tailored for soft robotics research, with pre-implemented modules for common external forces and actuation methods. 

In addition to mesh-based FEA approaches, hybrid particle–grid methods have recently emerged. A prominent example is ChainQueen~\cite{hu2019chainqueen}, which employs the Moving Least Squares Material Point Method (MLS-MPM). Like FEA, it is grounded in continuum mechanics, but it avoids mesh entanglement by representing material state on particles and transferring information through an Eulerian background grid. This allows it to naturally handle large deformations, self-collision, and even topological changes. However, ChainQueen relies on an explicit time-stepping scheme, which can be restrictive for the stiff systems of equations commonly encountered in soft robot dynamics.

To balance accuracy and efficiency, the third class of simulators adopts geometry-based reduced-order models. Elastica~\cite{Naughton2021_elastica_rl, Tekinalp2024} simulates Cosserat rods with bending, twisting, stretching, and shearing behavior and has been used effectively for one-dimensional soft robotic structures. However, its reliance on explicit time integration and its restriction to rod geometries limit its scalability and applicability to broader soft-body systems. SoRoSim~\cite{mathew2025reduced}, a MATLAB toolbox for soft-robot simulation, also employs the Cosserat-rod model and provides a modular interface for hybrid rigid–soft modeling, but likewise remains limited to one-dimensional rod elements.

DDG-based simulators take this approach further by discretizing geometric quantities directly, enabling efficient and accurate simulation of deformable structures. DisMech~\cite{choi2023dismech} introduced a C++ implementation of the Discrete Elastic Rods (DER) method for interconnected elastic rods with implicit time integration, improving both stability and performance over classical Cosserat models. However, it was limited to one-dimensional rod elements, supported only rod–rod contact, and did not include a generalized control framework. Building upon this foundation, MAT-DiSMech~\cite{lahoti2025matdismech} incorporated shell elements, enabling unified simulation of hybrid rod–shell systems. While this marked an important step toward general DDG-based modeling, its contact formulation remained restricted to rod–rod interactions, and the MATLAB environment imposed computational constraints. Moreover, the framework did not include an integrated control capability—highlighting the need for a more efficient and extensible platform.

In summary, while lumped-mass and FEA-based simulators excel respectively in speed and accuracy, geometry-based and DDG-informed methods provide a compelling middle ground, particularly for soft structures dominated by bending and twisting deformations. Existing tools, however, remain constrained by platform limitations, narrow modeling scope, or computational bottlenecks. To address these gaps, we introduce Py-DiSMech, a Python-native, high-performance DDG-based simulator that combines implicit integration with extensible formulations to robustly handle mixed rod–shell geometries and contact-rich dynamics. Py-DiSMech builds on well-established DDG models—the DER method for slender rods and discrete elastic-shell formulations from computer graphics for thin shells.

% 1. DER 2. DES
Since the introduction of the DER method by~\cite{bergou2008discrete}, this framework has seen widespread adoption for modeling rod-like soft robots and actuators~\cite{Huang2021, choi2023dismech, PneuNet}, deformable linear objects~\cite{choi_imc_2021, tong2023snap}, and digital hair~\cite{Goronowicz2015, Daviet2023}. Variants of this method have been successfully employed for motion planning and control in locomotion~\cite{Goldberg2019, Scott2021} and manipulation tasks~\cite{chen2024differentiable, 2024_Yu, chen2025deft}. DER has demonstrated accuracy comparable to finite element analysis (FEA) in capturing large deformations of soft rods in both static and dynamic settings.

Thin plates and shells, particularly for cloth modeling, have long been of interest in the computer graphics community. Among the various bending models, hinge-based methods—where bending energy is computed at edges shared by adjacent triangles—have been especially popular due to their simplicity~\cite{Baraff1998, Grinspun2003}. In our work, we implement two thin-shell bending models: a hinge-based formulation and another based on midedge normals~\cite{Grinspun2006}. The latter discretizes curvature operators directly and provides more robust dynamical behavior with respect to mesh quality, making it preferable for applications requiring high physical fidelity. This robustness, however, comes with added computational cost due to additional degrees of freedom associated with edges. These shell modeling techniques have also been leveraged for planning and controlling the manipulation of thin-shell objects~\cite{wang2024shellmanip}.

% Contact modeling
While accurate rod and shell models form the backbone of Py-DiSMech, an equally important component for realistic simulation is robust modeling of contact and friction. This remains a significant challenge due to the inherently discrete nature of contact events. Even finite element methods (FEA) often struggle to handle such interactions robustly~\cite{chrono_contact2017}. While commercial engines like NVIDIA PhysX support contact modeling, they often lack static friction capabilities, which are crucial for tasks such as grasping. Among recent methods, Incremental Potential Contact (IPC) and the Implicit Model for Contact (IMC) have gained popularity. IMC showed better computational speed than IPC at the cost of not enforcing no penetration~\cite{tong2023fully}; however, it was originally formulated for rod-like structures only. In our work, we have extended the IMC framework to handle shell and rod-shell contact scenarios, integrating it seamlessly into our simulation pipeline.

\section{Dynamical Modeling}
\label{sec: dynamics}

We discretize the structure into \( N \) nodes in 3D, connected by \( E \) edges (defining rods) and \( T \) triangles (forming shells). The user specifies:
\begin{itemize}
    \item an \( N \times 3 \) array of nodal positions,
    \item an \( E \times 2 \) array of edge node indices, and
    \item a \( T \times 3 \) array of triangle node indices.
\end{itemize}

\textit{Degrees of Freedom.}
Rod dynamics are modeled using the Discrete Elastic Rod (DER) framework~\cite{bergou2008discrete}, based on Kirchhoff rod theory. The degrees of freedom (DOFs) include nodal positions \( \mathbf{x} \) and edge twist angles \( \theta \), giving:
\begin{align}
\mathbf{q} = [\mathbf{x}_1, \dots, \mathbf{x}_N, \theta^1, \dots, \theta^E]^\top,
\end{align}
with total size \( 3N + E \).
For shells, we implement hinge-based~\cite{Grinspun2003} or mid-edge normal-based~\cite{Grinspun2006} bending models; the latter introduces additional DOFs \( \xi \) per shell-edge, extending the state vector to:
\begin{align}
\mathbf{q} = [\mathbf{x}_1, \dots, \mathbf{x}_N, \theta^1, \dots, \theta^E, \xi^1, \dots, \xi^Z]^\top,
\end{align}
where \( Z \) is the number of shell-edges.

\textit{Equations of Motion.}
The equations of motion are:
\begin{align}
\mathbf{M} \ddot{\mathbf{q}} = \mathbf{F}^{\text{elastic}} + \mathbf{F}^{\text{ext}} + \mathbf{F}^{\text{IMC}},
\end{align}
where \( \mathbf{M} \) is the diagonal lumped mass matrix; \( \mathbf{F}^{\text{elastic}} \), \( \mathbf{F}^{\text{ext}} \), and \( \mathbf{F}^{\text{IMC}} \) represent elastic, external, and self-contact/friction forces, respectively.

We solve the time evolution using implicit Euler integration. Introducing velocities \( \mathbf{u} = \dot{\mathbf{q}} \), the update equations are:
\begin{align}
    \mathbf{f} \equiv \mathbf{M} \frac{1}{\Delta t} \left( \frac{\mathbf{q}_{k+1} - \mathbf{q}_k}{\Delta t} - \mathbf{u}_k \right) +  \frac{\partial E^\textrm{elastic}_{k+1}}{\partial \mathbf q_\textrm{k+1}} &\nonumber \\
 - \mathbf{F}^{\text{IMC}}_{k+1} - \mathbf{F}^{\text{ext}}_{k+1} = \mathbf 0 &, 
    \label{eq:position_update}\\
    \mathbf{u}_{k+1} = \frac{\mathbf{q}_{k+1} - \mathbf{q}_k}{\Delta t}&,
    \label{eq:vel_update}
\end{align}
where subscript $k+1$ indicates evaluation of a quanity at $t=t_{k+1}$ and DOF $q (t_{k+1})$.
We use Newton-Raphson iterations to solve for \( \mathbf{q}_{k+1} \):
\begin{align}
\mathbf{q}_{k+1}^{(i+1)} = \mathbf{q}_{k+1}^{(i)} - \alpha\, \texttt{linearSolve}(\mathbf{J}, \mathbf{f}),
\end{align}
where \( \alpha \) is used to scale the magnitude of the step, it takes the value of $1$ by default and can be choosen adaptively using line search to aid in convergence, and \( \mathbf{J} \) is the Jacobian:
\begin{align}
\mathbf{J} = \frac{\mathbf{M}}{\Delta t^2} + \frac{\partial^2 E^{\text{elastic}}_{k+1}}{\partial \mathbf{q}_{k+1}^2} + \frac{\partial \mathbf{F}^{\text{IMC}}_{k+1}}{\partial \mathbf{q}_{k+1}} - \frac{\partial \mathbf{F}^{\text{ext}}_{k+1}}{\partial \mathbf{q}_{k+1}}.
\end{align}

Once convergence is achieved, velocities are updated using~\eqref{eq:vel_update}. Next, the section~\ref{sec:elasticity} details the formulations for elastic forces and section~\ref{sec:IMC_contact}, describes the contact model.

\subsection{Elastic Strains and Forces}
\label{sec:elasticity}

\textit{Stretching.}
Each stretching spring stores axial strain energy. The stretching strain for he $i$-th spring is given by,
\begin{align}
    \epsilon^\text{stretch}_i = \frac{\|\mathbf{e}^i\|}{\|\bar{\mathbf{e}}^i\|} - 1, 
\end{align}
where $\|\bar{\mathbf{e}}^i\|$ is the undeformed length of the $i$-th edge.

\begin{figure*}[t!]
\centerline{\includegraphics[clip, trim=0cm 6.9cm 0cm 0.1cm, width =\textwidth]{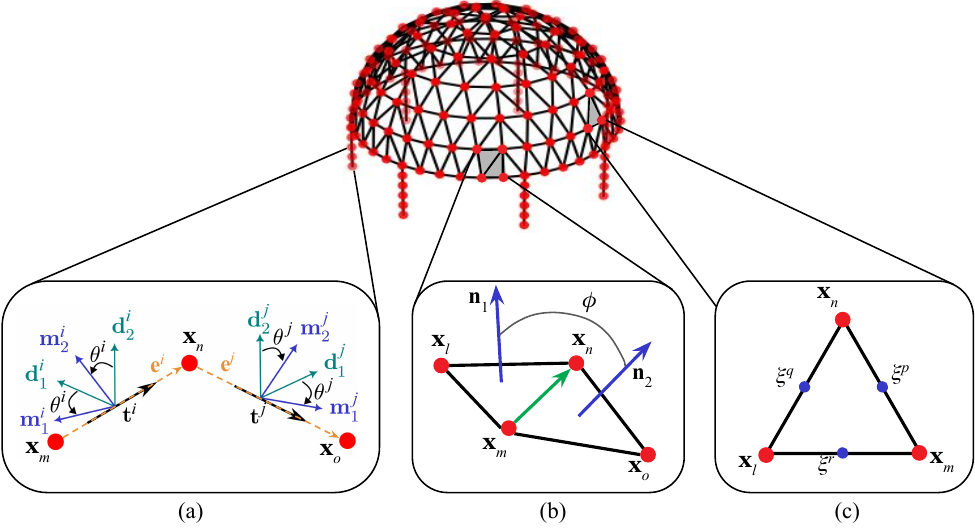}}
\caption{Schematics for the DDG-based dynamical models. (a) Stencil of the bending-twisting spring for an elastic rod, used for the DER algorithm. (b) Schematic of a hinge spring for the hinge-based bending model for a discrete elastic shell. (c) The discretization stencil for the mid-edge normal-based bending energy model. For the $i$-th shell-edge, $\xi^i$ is a scalar DOF that represents the rotation of the mid-edge normal about the edge.\cite{lahoti2025matdismech }}
\label{fig:schematics}
\end{figure*}

\textit{Bending and Twisting (Rods).}
Bending and twisting energy for slender rods is stored in bending-twisting springs. In Figure~\ref{fig:schematics}(b),  three nodes ($\mathbf x_m, \mathbf x_n,$ and $\mathbf x_o$) and two edges ($\mathbf e^i$ and $\mathbf e^j$) form a bending-twisting spring. The edge $\mathbf e^i$ (and $\mathbf e^j$) is the vector from $\mathbf x_m$ to $\mathbf x_n$ (and from $\mathbf x_n$ to $\mathbf x_o$).
Each rod-edge $\mathbf{e}^i$ has two sets of orthonormal frames, a reference frame $\{\mathbf{d}^i_1,\mathbf{d}^i_2,\mathbf{t}^i\}$ and a material frame $\{\mathbf{m}^i_1,\mathbf{m}^i_2,\mathbf{t}^i\}$. Both of these frames share the tangent vector along the edge $\mathbf{t}^i=\mathbf{e}^i/\|\mathbf{e}^i\|$ as one of the directors. The reference frame is initialized at $t = 0$ and updated at each timestep by parallel transport from the previous configuration—a key feature of Discrete Elastic Rods (DER) that enables high computational efficiency~\cite{bergou2010discrete}. The material frame is obtained by applying the twist angle $\theta^i$ about the shared tangent $\mathbf{t}^i$ to the reference frame.

Bending strain is measured at the center node $\mathbf{x_n}$ through the curvature binormal vector given by, 
\begin{align}
\label{equ:curvature}
    (\boldsymbol \kappa b)_k = \frac{\text{2}\mathbf{e}^{i}\times \mathbf{e}^j}{\|\mathbf{e}^{i}\|\|\mathbf{e}^{j}\|+\mathbf{e}^{i}\cdot \mathbf{e}^{j}}.
\end{align}
The scalar curvatures along the first and second material directors using the curvature binormal are 
\begin{align}
\label{equ:curvature_material}
    \kappa^{(1)}_k &= \frac{1}{2}(\mathbf{m}^{i}_2+\mathbf{m}^{j}_2)\cdot (\boldsymbol \kappa b)_k, \\
    \kappa^{(2)}_k &= -\frac{1}{2}(\mathbf{m}^{i}_1+\mathbf{m}^{j}_1)\cdot (\boldsymbol \kappa b)_k.
\end{align}
Using these scalar curvatures, the bending strain for the $k$-th bending-twisting spring is given by,
\begin{align}
\label{eq: bending_strain}
    \epsilon^\text{bend}_k = 
\begin{bmatrix}
\kappa^{(1)}_k - \bar{\kappa}^{(1)}_k \\
\kappa^{(2)}_k  - \bar{\kappa}^{(2)}_k
\end{bmatrix},
\end{align}
where $\bar{\kappa}^{(1)}_k$ and $\bar{\kappa}^{(2)}_k$ are the natural scalar curvatures.

The twisting strain between the two edges $i$ and $j$ corresponding to the $k$-th bending-twisting spring is 
\begin{align}
\label{equ:twist}
    \epsilon^\text{twist}_k = \theta^j - \theta^{i} + \Delta m_{k,\textrm{ref}},
\end{align}
where $\Delta m_{k,\textrm{ref}}$ is the reference twist, which is the twist of the reference frame as it moves from the $i$-th edge to the $j$-th edge.

An edge can be shared by multiple bending-twisting springs, especially when modeling rod networks. The formulation above assumes a specific edge orientation—$\mathbf{x}_m \to \mathbf{x}_n$ for the first edge and $\mathbf{x}_n \to \mathbf{x}_o$ for the second. As shown in Figure~\ref{fig:schematics}(g), for the bending-twisting spring ${\mathbf{x}_a, \mathbf{x}_b, \mathbf{x}_c; \mathbf{e}^i, \mathbf{e}^j}$, both rod edges $\mathbf{e}^i$ and $\mathbf{e}^j$ point toward node $\mathbf{x}_b$. To maintain consistency, the negative of $\mathbf{e}^j$ is used in calculations so that $\mathbf{e}^i$ points toward and $-\mathbf{e}^j$ points away from $\mathbf{x}_b$. Correspondingly, the reference frame vector $\mathbf{d}^j_1$, material frame vector $\mathbf{m}^j_1$, and rotation angle $\theta^j$ are multiplied by $-1$ when computing bending and twisting forces. Afterward, the resulting force vector and Jacobian are reoriented to their original configuration by multiplying the affected terms by $-1$ if such adjustments were applied. Note that alternatively flipping $\mathbf{e}^i$ instead of $\mathbf{e}^j$ would yield mathematically equivalent results.

\textit{Hinge Bending (Shells).}
Referring to Figure~\ref{fig:schematics}(c), the hinge spring comprises four nodes ($\mathbf x_l, \mathbf x_m, \mathbf x_n,$ and $\mathbf x_o$), and two of these nodes define the hinge, which in this case is the edge vector from $\mathbf x_m$ to $\mathbf x_n$. The hinge angle $\phi$ is defined as the angle between the normal vectors on these two shell triangles. 

The bending strain for the $i$-th hinge spring is defined as,
\begin{align}
    \epsilon^\text{hinge}_i = \phi_i - \bar{\phi}_i, 
\end{align}
where $\bar{\phi}_i$ is the natural hinge angle.

\begin{figure}[t!]
\centerline{\includegraphics[clip, trim=0cm 11.65cm 6.8cm 0cm, width =0.5\textwidth]{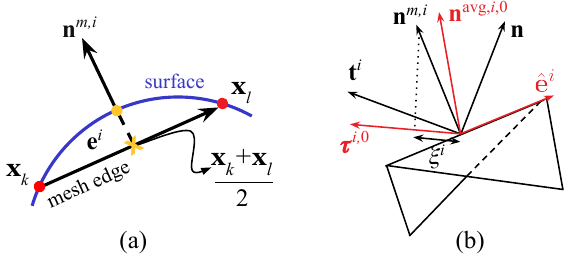}}
\caption{Schematics for midedge normal-based shell bending energy. (a) Definition of the mid-edge normal. The blue curve labeled ``surface'' denotes the actual surface of the shell being modeled, ``mesh edge'' $\mathbf{e}$ denotes the edge of a triangle in the mesh that approximates the surface. The mid-edge normal $\mathbf{n}^m$ for $\mathbf{e}$ is normal to the surface, which, when extrapolated, intersects the triangle edge at its midpoint. (b) The schematic showcasing the edge attached reference frame $\{\mathbf{n}^{\textrm{avg},i},\mathbf{\tau}^i, \hat{\mathbf{e}}^i$\} and other vectors used in the mid-edge normal bending method~\cite{Grinspun2006, lahoti2025matdismech }.}
\label{fig:midedge}
\end{figure}

\textit{Midedge Bending (Shells).}
In this method, the shape operator describes curvature and bending strains. To compute it discretely on a meshed surface, the mid-edge normal $\mathbf{n}^{m,i}$ is introduced: the smooth-surface normal intersecting edge $\mathbf{e}^i$ at its midpoint, as shown in Figure~\ref{fig:midedge}(a) (see~\cite{Grinspun2006}). As shown in Figure~\ref{fig:midedge}(b), each edge $\mathbf{e}^i$ has an attached frame \{$\mathbf{n}^{\textrm{avg},i}, \boldsymbol{\tau}^i, \hat{\mathbf{e}}^i$\}, where $\mathbf{n}^{\textrm{avg},i}$ is the average of the two adjacent face normals, $\hat{\mathbf{e}}^i$ is the unit edge vector, and $\boldsymbol{\tau}^i = \mathbf{n}^{\textrm{avg},i} \times \hat{\mathbf{e}}^i$. At the start of each timestep, we compute $\boldsymbol{\tau}^{i,0}$. For each edge, a scalar $\xi^i = \mathbf{n}^{m,i} \cdot \boldsymbol{\tau}^{i,0}$ represents the mid-edge normal’s rotation about the edge.

For a triangle with edges ${p, q, r}$ (Figure~\ref{fig:schematics}(d)), its shape operator is:
\begin{align}
\label{eq:lambda}
\Lambda_i = \sum_{k \in {p, q, r}} \frac{s^k \xi^k - (\mathbf{n}_i \cdot \boldsymbol{\tau}^{k,0})}{\bar{A}_i |\mathbf{\bar{e}}^k| (\hat{\mathbf{t}}^k \cdot \boldsymbol{\tau}^{k,0})} , \mathbf{t}^k \otimes \mathbf{t}^k,
\end{align}
where $\mathbf{n}_i$ is the triangle’s unit normal, $\bar{A}_i$ its undeformed area, $|\mathbf{\bar{e}}^k|$ the undeformed edge length, $\mathbf{t}^k$ the tangent perpendicular to the edge, $s^k \in {-1,1}$ accounts for normal ownership, and $\otimes$ denotes the outer product (Figure~\ref{fig:midedge}(b)).
The bending strain for the $i$-th triangle spring is then given by, 
\begin{equation}
\label{eq: midedge_strain}
    \epsilon^{\textrm{midedge}}_i = \sqrt{(1-\nu)\mathrm{Tr}((\Lambda_i-\bar{\Lambda}_i)^2) + 
\nu(\mathrm{Tr}(\Lambda_i) - \mathrm{Tr}(\bar{\Lambda}_i) )^2},
\end{equation}
where $\nu$ denotes the Poisson's ratio  of the material, $\bar{\Lambda}_i$ denotes the shape operator of the $i$-th triangle in the undeformed or natural configuration, and $\mathrm{Tr}(\,)$ denotes the trace of a matrix. Note that this definition is not purely geometric but has some material specific property due to the use of Poisson's ratio for scaling, this is since we are actually adding up two different forms of strain expressions, the collective term is actually a linear combination of the two strains $\mathrm{Tr}((\Lambda_i-\bar{\Lambda}_i)^2)$ and $(\mathrm{Tr}(\Lambda_i) - \mathrm{Tr}(\bar{\Lambda}_i) )^2$ which are purely geometric. The collective term is still dimensionless and we call it ``strain" for consistent notation for the energy expressions later in \eqref{eq: elastic_energy}.

\textit{Rod-Shell Joint.}
When a node is shared between a rod edge and one or more shell triangles (a \textit{joint node}), all associated deformations are combined. Shell edges near the joint are treated as rod edges, including material frames and twist DOFs, and bending-twisting springs are assigned across all valid three-node, two-edge combinations. For mid-edge shell models, joint edges carry both \(\theta\) and \(\xi\) DOFs. The joint node thus experiences combined forces from stretching, rod bending/twisting, and shell bending.

For the $i$-th spring, the elastic strain energy of deformation type ``def” (e.g., stretch, bend, twist, hinge, or midedge) is computed using the strains defined above and the corresponding force and Jacobian are then obtained via the chain rule of differentiation as follows,
\begin{align}
\label{eq: elastic_energy}
    E^\text{def}_i &= \frac{1}{2} K^\text{def}_i ((\epsilon^\text{def}_i)^T\epsilon^\text{def}_i),\\
    \label{eq: elastic_force}
    \mathbf{F}^\text{def}_i &= -\nabla_\mathbf{q} E^\text{def}_i = -K^\text{def}_i \epsilon_i^\text{def}  \nabla_\mathbf{q} \epsilon_i^\text{def} \\
    \nonumber
    \mathbf{J}^\text{def}_i &= -\nabla^2_\mathbf{q} E^\text{def}_i \\ 
    \label{eq: elastic_jacob}
    &= -K^\text{def}_i \left[ \epsilon_i^\text{def}  \nabla^2_\mathbf{q} \epsilon_i^\text{def}
    + (\nabla_\mathbf{q} \epsilon_i^\text{def})^2 \right]
\end{align}
Here $K^\text{def}_i$ is the stiffness of the $i$-th spring for ``def" type of strain energy; the expression for $K$ for the different type of strain energies are given in Table \ref{tab:stiffness_expressions}. Here, $E^\text{rod}$ and $E^\text{shell}$ denote the Young's moduli for rod and shell respectively; $G^\text{rod}$ denotes the shear modulus for the rod material ($G = E/(2(1+\nu^\text{rod})$, where $\nu^\text{rod}$ is the Poisson's ratio for the rod material); $A$ denotes the area of cross-section of the rod; $I_1$ and $I_2$ denote the area moment of inertia for the rod cross-section along the two in-plane perpendicular directions; $J$ denotes the polar moment of inertia for the rod cross-section; $h$ denotes the thickness of the shell and $\nu^\text{shell}$ denotes the Poisson's ratio for the shell.
\begin{table}[!t]
\caption{Stiffness expressions for various deformation modes in rods and shells.}
\centering
\renewcommand{\arraystretch}{1.4}
\begin{tabular}{|l l|}
% \hline
% \textbf{Stiffness Type} & \textbf{Expression} \\
\hline
$K^\text{stretch, rod}$ & $E^\text{rod} A$ \\
% \hline
$K^\text{stretch, shell}$ & $\frac{\sqrt{3}}{4} E^\text{shell} h \|\bar{\mathbf{e}}^{i}\|$ \\
% \hline
$K^\text{bend}$ & $\begin{bmatrix}
E^\text{rod} I_1 & 0 \\
0 & E^\text{rod} I_2
\end{bmatrix}$ \\
% \hline
$K^\text{twist}$ & $G^\text{rod} J$ \\
% \hline
$K^\text{hinge}$ & $\frac{1}{\sqrt{3}} \frac{E^\text{shell} h^3}{12}$ \\
% \hline
$K^\text{midedge}$ & $\frac{E^\text{shell} h^3}{24(1 - (\nu^\text{shell})^2)}$ \\
\hline
\end{tabular}
\label{tab:stiffness_expressions}
\end{table}

Note that in case of type ``bend", the strain $\epsilon^\text{bend}$ \eqref{eq: bending_strain} is a vector of two elements, the stiffness $K$ is a diagonal matrix of size $2\times2$, and the two diagonal values can be different depending on the cross-section of the rod. Also note that since the expression for strain for midedge bending are fairly complex, we have directly computed the derivatives of the energy with respect to the DOF vector instead of using the above chain rule through strains to avoid square rooting operations.

The total elastic energy combines contributions from discrete spring elements:
\begin{align}
% E^{\text{elastic}} = \sum_i E_i^{\text{stretch}} + \sum_i E_i^\text{bend} + \sum_i E_i^{\text{twist}} + \sum_i E_i^{\text{shell}},
E^{\text{elastic}} = \sum_{\text{def}} \sum_{j \in \text{springs}} E_j^{\text{def}},
\end{align}
where def $\in$ \{stretch, bend, twist, hinge/midedge\}. The same applies to the computation of the total elastic force and Jacobian.

\subsection{Self-contact and Friction}
\label{sec:IMC_contact}
\begin{figure}[t!]
\centerline{\includegraphics[clip, trim=0cm 4.5cm 0cm 0cm, width=1\columnwidth]{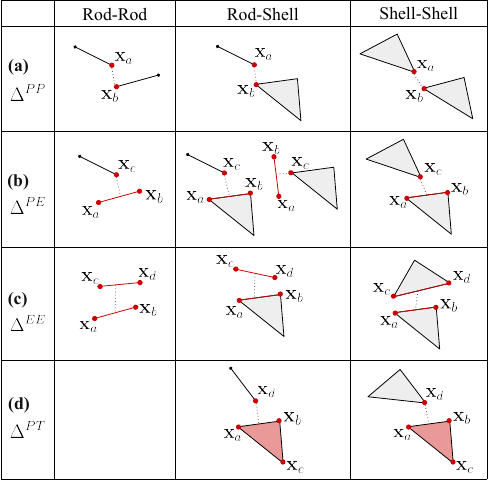}}
\caption{Types of contact interactions modeled in Py-DiSMech. (a) Point-to-Point, (b) Point-to-Edge, (c) Edge-to-Edge, and (d) Point-to-Triangle. Contact types ((a)–(c)) can occur in all contact-pair configurations—Rod-Rod, Rod-Shell, and Shell-Shell, whereas Point-to-Triangle contact (d) arises only in Rod-Shell and Shell-Shell interactions.}
\label{fig:contact}
\end{figure}

Similar to how elastic energies are evaluated per spring, contact energy and forces are computed per contact pair. Each contact pair may exhibit one of four contact types: (i) Point-to-Point, (ii) Point-to-Edge, (iii) Edge-to-Edge, or (iv) Point-to-Triangle as shown in Figure \ref{fig:contact}. Note that Edge-to-Triangle contact arises only under interpenetration, and properly tuned contact stiffness can prevent such cases. Nevertheless, since our penalty energy method doesn't strictly enforce no penetrations, we ensure that if such a situation is indeed encountered, it is dealt with within the simulator.
The contact type for each pair is determined in a batched manner. For edge-edge interactions, we employ Lumelsky's algorithm~\cite{Lumelsky1985}; for triangle-triangle interactions, we use a batched version of the barycentric coordinates computation method inspired by the NVIDIA PhysX solver~\cite{nvidia2024_physx_release}.
% \begin{figure*}[t!]
% \centerline{\includegraphics[clip, trim=0cm 0cm 0cm 0cm, width =0.8\textwidth]{Figures/Contact_schematics.pdf}}
% \caption{Types of contact}
% \label{fig:contact}
% \end{figure*}

The distance $\Delta$ between the entities of the contact pairs is computed in a batched manner using the analytical functions corresponding to the type of contact. If the contact type is Point-to-Point, and the nodes $\mathbf{x}_a$ and $\mathbf{x}_b$ are the ones in close proximity as shown in Figure \ref{fig:contact} (a), 
\begin{align}
    \Delta^{PP} = ||\mathbf{x}_a - \mathbf{x}_b||.
\end{align}
Else if the contact type is Point-to-Edge, and the node $\mathbf{x}_c$ corresponds to the ``point" and nodes $\mathbf{x}_a$ and $\mathbf{x}_b$ form the ``edge" as shown in Figure \ref{fig:contact} (b), then
\begin{align}
    \Delta^{PE} = \frac{||(\mathbf{x}_a - \mathbf{x}_b) \times (\mathbf{x_b}-\mathbf{x_c})||}{||\mathbf{x}_a - \mathbf{x}_b||}.
\end{align}
Else if the contact type is Edge-to-Edge, and the nodes $\mathbf{x}_a$ and $\mathbf{x}_b$ form one of the two edges and the nodes $\mathbf{x}_c$ and $\mathbf{x}_d$ form the other edge as shown in Figure \ref{fig:contact} (c), then
\begin{align}
    \Delta^{EE} = |(\mathbf{x}_a - \mathbf{x}_c) \cdot \frac{(\mathbf{x}_a - \mathbf{x}_b) \times (\mathbf{x}_c-\mathbf{x}_d)}{||(\mathbf{x}_a - \mathbf{x}_b) \times (\mathbf{x}_c-\mathbf{x}_d)||}|.
\end{align}
Finally, if the contact type is Point-to-Triangle, and the node $\mathbf{x}_d$ corresponds to the ``point" and the nodes $\mathbf{x}_a$, $\mathbf{x}_b$ $\mathbf{x}_c$ form the ``triangle" edges as shown in Figure \ref{fig:contact} (d), then
\begin{align}
    \Delta^{PT} = |(\mathbf{x}_d - \mathbf{x}_a) \cdot \frac{(\mathbf{x}_b - \mathbf{x}_a) \times (\mathbf{x_c}-\mathbf{x}_a)}{||(\mathbf{x}_b - \mathbf{x}_a) \times (\mathbf{x_c}-\mathbf{x}_a)||}|.
\end{align}

For each contact pair, $\Delta$ distance apart, we compute the contact energy as follows,
\begin{align}
E^{\text{con}} =
\begin{cases}
(2h - \Delta)^2 & \Delta \leq 2h - \delta, \\
0 & \Delta \geq 2h + \delta, \\
\left(\frac{1}{K_1} \log \big(1 + e^{K_1 (2h - \Delta)} \big) \right)^2 & \text{otherwise},
\end{cases}
\end{align}
where \( K_1 = 15 / \delta \) is a stiffness parameter, and \( \delta \) is the user-defined contact distance tolerance. 

The contact force on node \( i \) of the contact pair is:
\begin{align}
\mathbf{F}_i^{\text{con}} = -\frac{\partial E^{\text{con}}}{\partial \Delta} \frac{\partial \Delta}{\partial \mathbf{q}_i}.
\end{align}

The Coulomb friction force is then given by,
\begin{align}
\mathbf{F}_i^{\text{fr}} = - \mu \gamma \hat{\mathbf{u}} \| \mathbf{F}_i^{\text{con}} \|, \quad
\gamma = \frac{2}{1 + e^{-K_2 \| \mathbf{u} \|}} - 1,
\end{align}
where \( \mu \) is the friction coefficient, \( \mathbf{u} \) is the tangential relative velocity between entities that comprise the contact pair, \( K_2 = 15 / \nu \), and \( \nu \) is the user-defined slipping tolerance.

The total contact-friction force sums over all contact pairs:
\begin{align}
\mathbf{F}^{\text{IMC}} = \sum_{\{a, b, c, d\} \in \text{pairs}} \sum_{i \in \{a, b, c, d\}} \big( \mathbf{F}_i^{\text{con}} + \mathbf{F}_i^{\text{fr}} \big).
\end{align}

For gradient and Jacobian expressions, we refer readers to ~\cite{tong2023fully}.

\section{Software Structure and Usage}
\label{sec: software}

 \begin{figure*}[t!]
\centerline{\includegraphics[width =\textwidth]{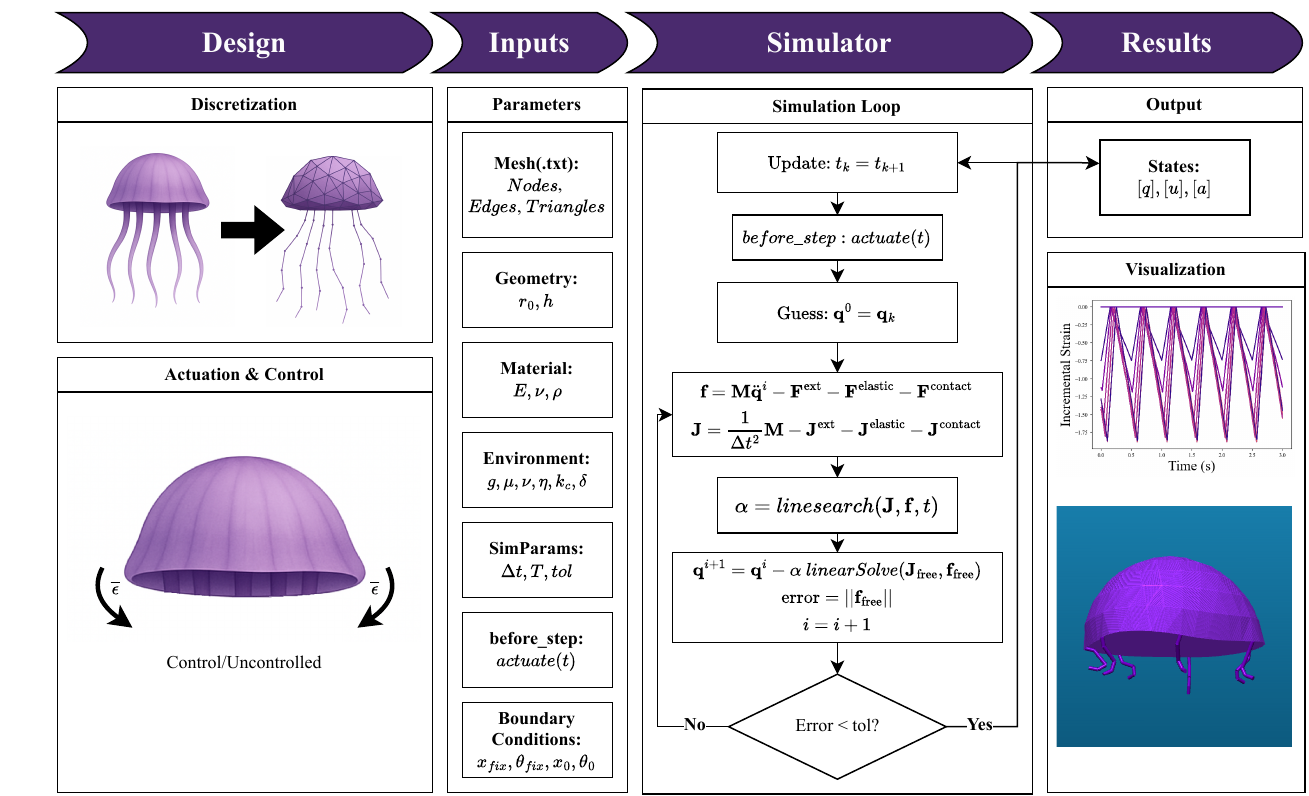}}
\caption{High-level flowchart of the Py-DiSMech workflow illustrating how design inputs are processed through the simulation to generate outputs that inform soft-robot design decisions.}
\label{fig::flowchart}
\end{figure*}

Simulating the motion of a soft robot requires a comprehensive description of its physical and computational characteristics. The inputs include:
\begin{itemize}
    \item \textbf{Geometry:} The mesh representation of the soft robot, which includes nodal coordinates and connectivity, and cross-sectional dimensions.
    \item \textbf{Material properties:} Density, Young’s modulus, and Poisson’s ratio, which define the inertia and constitutive behavior of the material.
    \item \textbf{Boundary conditions:} Specification of fixed and free nodes or edges, which constrain motion and define the robot’s interaction with its supports or actuation interfaces.
    \item \textbf{External environment:} The kind of external forces acting on the robot, such as gravity, contact, or fluid drag.
    \item \textbf{Simulation parameters:} Numerical settings including the time step, total simulation time, and convergence tolerances.
    \item \textbf{Actuation:} If intending to actively cause motion, prescribed inputs such as boundary displacements or natural strain fields are required.
\end{itemize}
All the above information is provided by the user as input to the simulator.

The simulation process begins by initializing a SoftRobot class instance, which encapsulates the robot’s mesh, geometry, material parameters, simulation settings, and environmental information. Once instantiated, boundary conditions are applied to define fixed and free degrees of freedom of the robot. Subsequently, a TimeStepper class is initialized. This module orchestrates the temporal integration of the equations of motion and, during initialization, automatically constructs the relevant energy and force models. These include elastic energy classes (for any or all of the stretching, bending and twisting deformations), contact energy classes (for self-collision and optionally, friction), and external force classes (such as gravity or hydrodynamic drag). If the simulation involves active actuation—as opposed to a passive dynamic response—custom actuation subroutines are defined. These routines typically modify boundary conditions or natural strain parameters at each time step and are invoked automatically by the time-stepping loop.

With the system fully defined, the user calls the simulate() method of the TimeStepper. This routine advances the system in time using an implicit integration scheme, accounting for all forces, constraints, and actuation effects. The solver logs the robot’s degrees of freedom (positions, velocities, and strains) at user-specified intervals for analysis or post-processing. Visualization and analysis tools are integrated within the simulator, allowing users to render the robot’s motion, inspect deformation fields, and plot quantities of interest such as energy evolution or tip trajectories. Figure~\ref{fig::flowchart} shows the overall steps in the simulation framework.

% \begin{figure*}[t!]
% \centerline{\includegraphics[clip, trim=0cm 0.5cm 0cm 0.5cm, width =\textwidth]{Figures/computational_graph.pdf}}
% \caption{High-level software structure represented as a computational graph, divided into setup and simulation relationships. Most classes are designed for use during setup to minimize runtime overhead.}
% \label{fig::code_graph}
% \end{figure*}

\begin{figure*}[t!]
\centerline{\includegraphics[clip, trim=0cm 0.5cm 0cm 0.5cm, width =\textwidth]{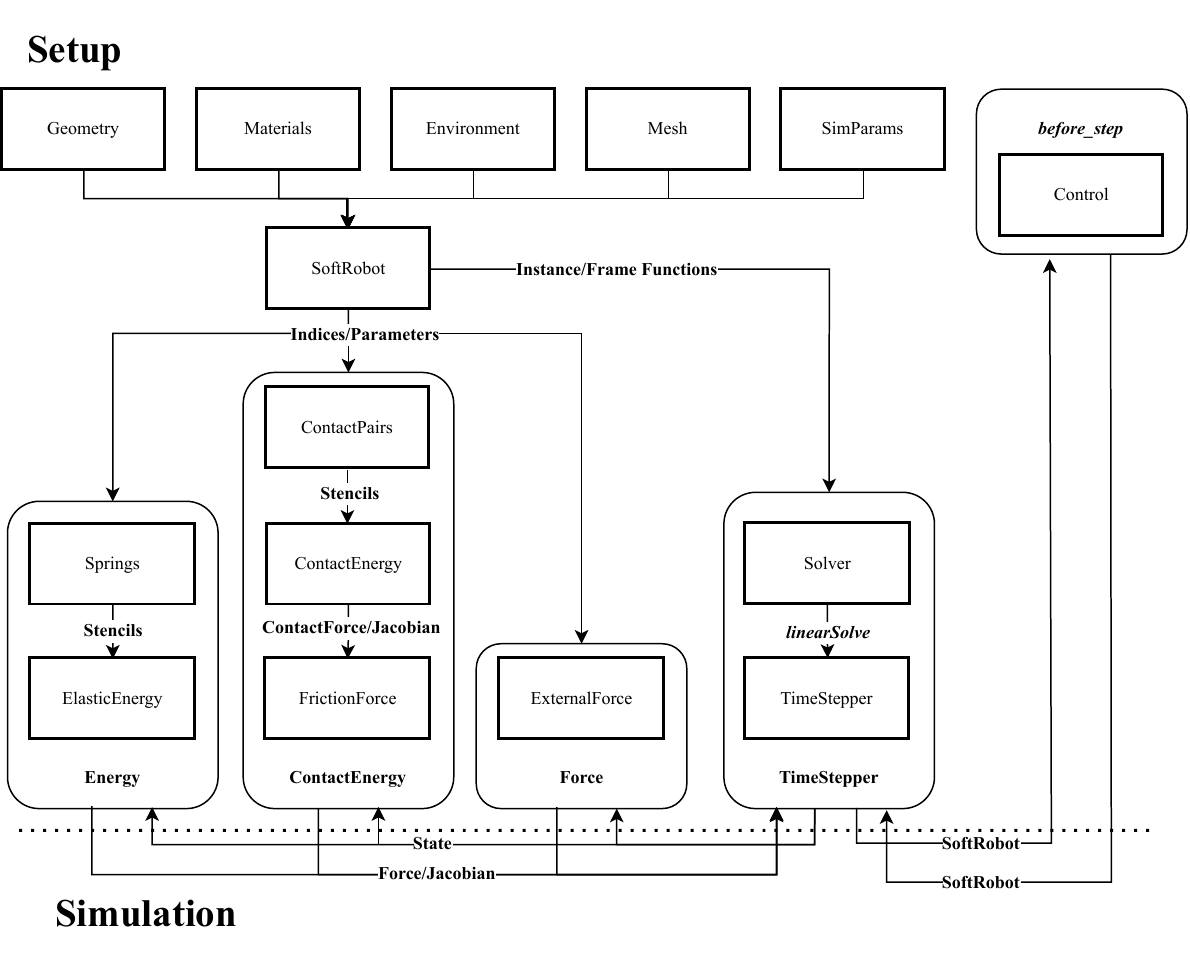}}
\caption{High-level software structure represented as a computational graph, divided into setup and simulation relationships. Most classes are designed for use during setup to minimize runtime overhead.}
\label{fig::code_graph}
\end{figure*}

% \subsection{Software Structure}

Our simulator is divided into classes that are used before simulation during setup, during simulation, or in both phases. For the setup phase, we have the classes Geometry, Materials, Environment, Mesh, and SimParams, which are used to initialize a SoftRobot object. In simulation, the TimeStepper, Force, and Energy subclasses act upon this SoftRobot object and accompanying State objects to generate a trajectory of states. In the subsequent subsections, we will describe the key implementation details of the simulator components.

\subsubsection{SoftRobot}
To initialize a SoftRobot object, the user must first instantiate five helper objects that serve as configuration: Geometry, Mesh, Material, SimParams, and Environment. The ``Geometry'' object constitutes the cross-sectional dimensions and ``Mesh'' object contains the nodal coordinates and their connectivity; together these two objects fully define the complete geometric configuration of the structure. ``Material'' object constitutes the material properties, namely density, Young's modulus, and Poisson ratio. ``SimParams'' include the parameters for time stepping---step size, total simulation time, tolerances, and various flags that are used to enforce things like static simulation, 2D simulation, which dynamical model to use for the shell simulation and so on. Once the SoftRobot object is created, users can set boundary conditions by calling \textit{robot.fix\_nodes} for nodes, and \textit{robot.fix\_edges} for edges. Additionally, users can also set initial conditions for nodes and edges using \textit{robot.move\_nodes} and \textit{robot.twist\_edges} respectively. These functions can also be invoked within the \textit{before\_step} actuation function to mimic a moving boundary condition-based actuation.

Each SoftRobot object holds a reference to a frozen dataclass called RobotState, which contains the degree of freedom vector $\mathbf{q}$, velocity vector $\mathbf{u}$, acceleration $\mathbf{a}$ vector, reference frames \{$\mathbf{d}_1, \mathbf{d}_2$\}, and material frames \{$\mathbf{m}_1, \mathbf{m}_2$\}, at a particular time step. To index a specific node or edge within these flattened vectors, \textit{robot.map\_node\_to\_dof} and \textit{robot.map\_edge\_to\_dof} return indexing arrays that select the degrees of freedom associated with the requested features. Unlike other classes, SoftRobot can be flexibly interfaced during both setup and simulation via the \textit{before\_step} callback.

\subsubsection{Springs}
In our discrete differential geometry-based elastic energy formulation, the energy is computed by summing the contributions from stencils across a larger structure. This approach efficiently maps to computational resources, allowing for non-divergent parallelization on GPUs and vectorized operations on CPUs. For standard Discrete Elastic Rod energy formulations, we automatically determine the two-node and three-node stencils used for stretching, and for bending and twisting, respectively. If a downstream user wants to implement a custom elastic energy formulation using these stencils, they can easily access them via \textit{robot.stretch\_springs} and \textit{robot.bend\_springs}, \textit{robot.twist\_springs}. For shells, we provide both \textit{robot.hinge\_springs} and \textit{robot.triangle\_springs}, used for hinge-based and mid-edge bending, respectively. The contributions from the springs to the elastic forces and Jacobian are computed in one go through vectorization, which helps reduce the computational cost significantly.

With each spring, there are associated \textit{nat\_strain} and \textit{inc\_strain} properties which are primarily used to implement a changing natural strain-based actuation strategy, hence these properties are made modifiable and are usually configured if needed through the \textit{before\_step} function. 

\subsubsection{ElasticEnergy}
To generalize the implementation of elastic energy formulations, we created an ElasticEnergy base class that abstracts away common chain rule computations and standardizes the interface. Specifically, we decouple the energy and strain calculations so that for each different kind of elastic energy, i.e. Stretching, Bending, Twisting, etc., the derived energy class implements the strain and derivatives of strain computations, while the Energy and its derivatives which in turn depend on the strain computations, are computed using the same function for each of the derived energy classes. For example, the StretchEnergy class derived from the ElasticEnergy base class uses a two-node stencil called StretchSprings and defines the strain functions to compute longitudinal strain and its gradient, and hessian. This approach minimizes the effort required to extend the simulator, for example, by adding a new kind of deformation such as shearing, since users only need to define a new stencil and implement \textit{get\_strain}, \textit{grad\_hess\_strain}. It also makes it straightforward to define energy equations with arbitrary strain contributions, such as Sadowsky's Ribbon~\cite{sadowsky_ribbon}.

% Within the simulator, the public method grad\_hess\_energy is repeatedly called within the inner loop using a properly modified state value. This design allows the TimeStepper to support various integration techniques, where the evaluated state may differ from the initial or final state.

\subsubsection{ContactPairs}
To represent contact interactions, we introduce the concept of contact pairs, which carry the contact energy, analogous to springs in the elastic energy formulation.
\begin{itemize}
    \item For \textbf{rod-rod contact}, contact pairs consist of nodes from non-adjacent edges.  
    \item For \textbf{rod-shell contact}, contact pairs are defined between nodes of a non-adjacent edge and a triangle.  
    \item For \textbf{shell-shell contact}, contact pairs involve nodes from non-adjacent triangles.
\end{itemize}
These configurations are illustrated in Figure~\ref{fig::first}(b). Similar to the springs, the contributions from the ContactPairs to the contact and friction forces and Jacobian are computed in one go through vectorization.

\subsubsection{ContactEnergy}
We model self-contact using a penalty energy method, following the approach of~\cite{tong2023fully}. To enable efficient computation, we define a candidate set at the beginning of each timestep during the first Newton-Raphson iteration. This set includes contact pairs that are likely to interact during the timestep, allowing us to limit collision checks to this subset and thereby reduce computational cost. Similar to ElasticEnergy class, the ContactEnergy base class decouples the energy and separation calculations so that for each different kind of contact energy, i.e. Rod-Rod, Rod-Shell, Shell-Shell, the specific derived energy class implements the computation of separation and its derivatives, while the Energy and its derivatives, are computed using the same function for each of the derived energy classes.

\subsubsection{FrictionForce}
Friction force is also defined using ContactPairs. It uses the values of separation and contact force and jacobian from the ContactEnergy classes. Unlike the Contact Energy, we do not associate an energy for Frictional force, since it is a non-conservative force and hence doesn't have the notion of Energy. Instead, we define the frictional force as proportional to the normal component of the contact force in the direction opposite to the relative velocity of the contact pair and compute the frictional Jacobian analytically using chain rule.

\subsubsection{ExternalForce}
External forces are incredibly varied and so this base class is provided the most amount of information in our simulator. Like other energy and force classes, an ExternalForce subclass implements \textit{compute\_force} and \textit{compute\_force\_jacobian}, where they are provided a SoftRobot object for mass and connectivity and an evaluation state. Given the lax requirements, duck typing these functions is intended as external forces are unlikely to share common implementation.

\subsubsection{TimeStepper}
Besides SoftRobot, the primary class simulation users will interact with is the TimeStepper. This class encapsulates all supporting method calls and data augmentation within solver iterations. As many different integration schemes are used for different experiments, we decided to implement TimeStepper as an abstract base class, and different timestepping methods are implemented as derived subclasses. We provide the implementations for ImplicitEulerTimeStepper, ImplicitMidPointStepper, and NewmarkBetaTimeStepper. These mainly vary in the inertial force calculations and how position, velocity, and acceleration are updated at the end of a successful step. From a user's perspective, they will initialize a TimeStepper object by passing a SoftRobot object which contains SimParams object. Initializing TimeStepper object initializes the standard ElasticEnergy, ContactEnergy, and ExternalForce subclasses as required by the Environment or SoftRobot. Custom energy and force classes can be easily integrated into an existing TimeStepper object.
Users can add their actuation subroutines to be executed before every timestep by assigning \textit{stepper.before\_step} to any callable that takes a SoftRobot object and a float for the time in seconds. Specifically, the user is provided a functional hook where they can implement the function $f(\textrm{SoftRobot}, t) \rightarrow \textrm{SoftRobot}$.

To standardize potential force and Jacobian contributions, all base classes implement a \textit{get\_grad\_hess(state)} method, which is called by the inner TimeStepper loop. The base classes ElasticEnergy and ContactEnergy contain commonly used chain rule computations, following the ``Don't Repeat Yourself" principle.

\subsubsection{Solver}

To handle both dense and sparse problems, we modularized our \textit{linearSolve} to enable interchangeable solving methodologies. In our base framework, we offer a standard Numpy dense pseudoinverse and a sparse solver through PyPardiso\cite{pypardiso}. To support both methods, we offer both dense and sparse matrix accumulation to avoid excess overheads.

\subsubsection{Control}
The Control class provides a modular interface for implementing feedback control-based actuation strategies. It defines an \textit{update()} method, called through the \textit{before\_step} callback, which adjusts the natural strains in response to the error between desired and measured deformation states. A built-in PI controller enables curvature- and stretch-based actuation by continuously updating the natural strain fields to track prescribed reference trajectories. The design is easily extensible—users can adapt the \textit{update()} logic to control other deformation modes (e.g., twist or hinge-bending) or to implement alternative feedback strategies such as model-predictive control without modifying the core simulation architecture.
\begin{table*}[t!]
\caption{Comparison of the computational time between PyElastica and Py-DiSMech for six simulation experiments: (i-ii) Helix under gravity, (iii-v) Cantilever under gravity, (vi) Snake in viscous fluid}
\centering
\renewcommand{\arraystretch}{1.2}
\begin{tabularx}{\textwidth}{>{\centering\arraybackslash}X
                                >{\centering\arraybackslash}X
                                >{\centering\arraybackslash}X
                                >{\centering\arraybackslash}X 
                                >{\centering\arraybackslash}X
                                >{\centering\arraybackslash}X}
\toprule
\textbf{Plot ID} & \textbf{Experiment} & \textbf{Integrator} & \textbf{E (Pa)} & \textbf{Wall Clock Time (s)} & \textbf{Time Step (s)} \\
\midrule
\multicolumn{6}{c}{\textit{PyElastica}} \\
\midrule
 & Cantilever & Verlet         & 1e5 & 2.51   & 9e-5 \\
a & Cantilever & Verlet        & 1e6 & 7.40   & 3e-5 \\
 & Cantilever & Verlet         & 1e7 & 24.58  & 9e-6 \\
b & Snake      & Verlet         & 1e6 & 8.39   & 3e-5 \\
c & Helix      & Verlet         & 1e7 & 5.61   & 4e-5 \\
d & Helix      & Verlet         & 1e9 & 229.52 & 1e-6 \\
\midrule
\multicolumn{6}{c}{\textit{Py-DiSMech}} \\
\midrule
 & Cantilever & Newmark-Beta    & 1e5 & 2.41 & 1e-2 \\
 & Cantilever & Implicit Euler  & 1e5 & 1.27 & 1e-2 \\
a & Cantilever & Newmark-Beta   & 1e6 & 4.77 & 1e-2 \\
 & Cantilever & Implicit Euler  & 1e6 & 2.37 & 1e-2 \\
 & Cantilever & Newmark-Beta    & 1e7 & 3.83 & 1e-2 \\
 & Cantilever & Implicit Euler  & 1e7 & 2.35 & 1e-2 \\
b & Snake      & Newmark-Beta   & 1e6 & 4.13 & 1e-2 \\
 & Snake      & Implicit Euler  & 1e6 & 4.3  & 1e-2 \\
c & Helix      & Newmark-Beta   & 1e7 & 5.33 & 5e-2 \\
 & Helix      & Implicit Euler  & 1e7 & 0.79 & 5e-2 \\
d & Helix      & Newmark-Beta   & 1e9 & 20.94 & 1e-2 \\
 & Helix      & Implicit Euler  & 1e9 & 7.16  & 1e-2 \\
\bottomrule
\end{tabularx}
\label{tab:pyelastica_v_dismech}
\end{table*}

%%%%%%%%%%%%%%%%%%%%%%%%%%%%%%%%%%

 \begin{figure*}[t!]
\centerline{\includegraphics[width =\textwidth]{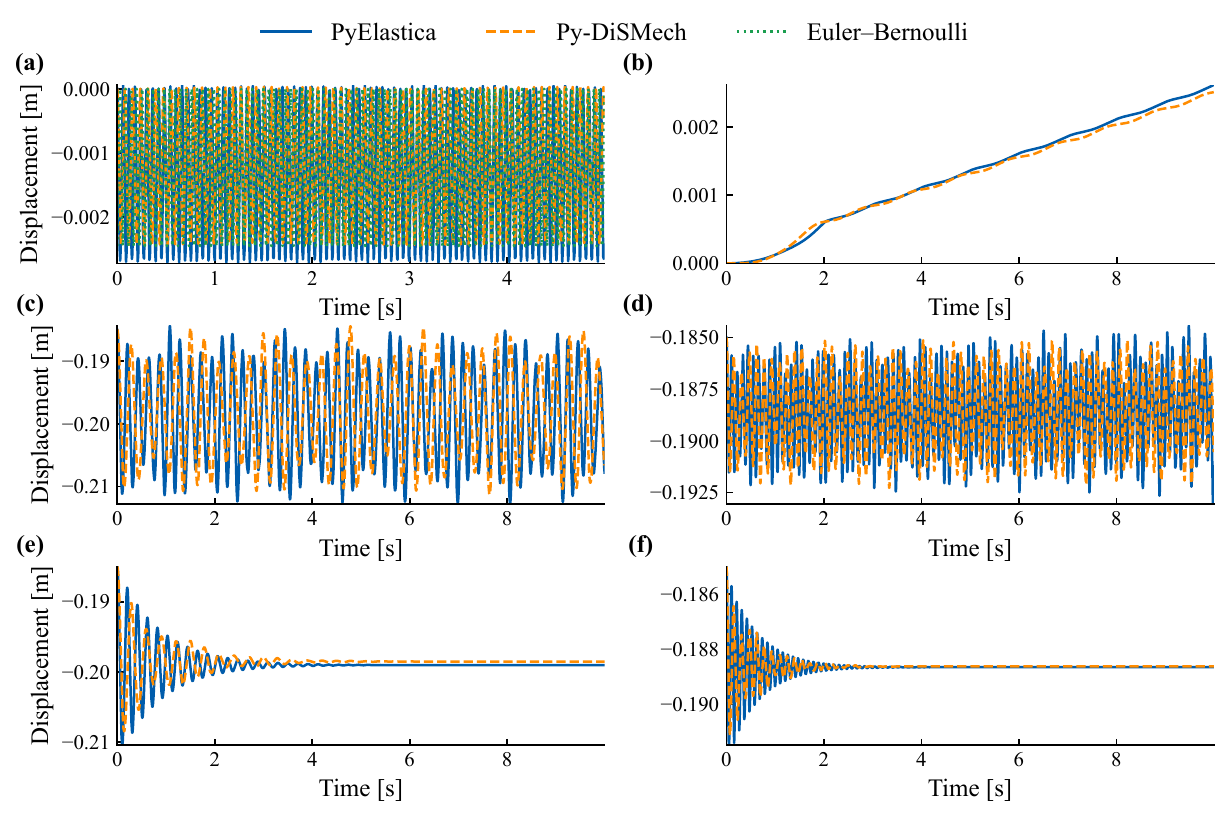}}
\caption{Comparison of Py-DiSMech, PyElastica, and theory (when applicable). (a) Undamped cantilever 1 MPa. (b) Actuated snake forward locomotion. (c) Undamped helix 10 MPa. (d) Undamped helix 1 GPa. (e) 10 MPa helix dampened for resting position. (f) 1 GPa helix dampened for resting position.}
\label{fig:comaprison_table}
\end{figure*}
%%%%%%%%%%%%
Figure \ref{fig::code_graph} illustrates the overall software structure, showing how setup and simulation modules interact through a unified data flow centered on the \textit{SoftRobot} object. This modular organization and vectorized implementation render Py-DiSMech highly scalable in both functionality and performance. New physical models, numerical solvers, or control strategies can be incorporated as independent components, and the underlying vectorized routines enable efficient computation even for large, high-resolution meshes.

\section{Comparison With State-of-the-Art}
\label{sec: comparison}

To contextualize our simulator against existing state-of-the-art offerings, we compared the simulation time and accuracy of cantilever (Figure ~\ref{fig:comaprison_table}~(a)) and helical rod (Figure ~\ref{fig:comaprison_table}~(c,d)) experiments under gravity ($9.81$~m/s$^2$) using Py-DiSMech and PyElastica~\cite{gazzola2018}. Both frameworks share the same programming language and discretization philosophy but differ in their dynamical modeling and time-integration schemes. A simple cantilevered rod was chosen for its analytical reference solution, while a helical rod was used to test performance under geometric and numerical complexity. An actuated snake locomoting through a viscous fluid was selected as the final case to assess the impact of actuation on performance. For each experiment in Figure~\ref{fig:comaprison_table}, the reported wall-clock times represent the average of five runs on an AMD Ryzen~9~9950X CPU with 64~GB of DRAM.

To ensure a fair comparison, we conducted all experiments over a range of stiffness values and geometries (10~MPa to 1~GPa), highlighting the trade-offs between implicit and explicit integration and identifying regimes where each method performs best. When comparing computational times, we evaluated PyElastica against both the Implicit Euler and Newmark–Beta integration schemes in Py-DiSMech to avoid bias introduced by the inherent damping in the Implicit Euler method.

In the cantilever experiments (Figure ~\ref{fig:comaprison_table}~(a)), damping was applied to facilitate convergence to a static equilibrium, allowing quantitative accuracy comparisons. Each rod consisted of 101 nodes, $\rho = 1000~\text{kg/m}^3$, radius $r = 0.02$~m, and length $0.1$~m, with varying Young’s modulus. Since Py-DiSMech and PyElastica implement damping differently, we experimentally tuned Py-DiSMech’s $\eta$ parameter to match PyElastica’s energy decay rate. Accuracy was assessed by comparing the final tip displacement to the analytical Euler–Bernoulli beam solution, as shown in Figure ~\ref{fig:comaprison_table}~(a). As stiffness increases, the governing ODE system becomes increasingly stiff, requiring smaller time steps for explicit solvers to maintain stability. In contrast, implicit solvers accommodate these stiff equations with larger time steps while preserving accuracy. Using Implicit Euler, we observed a 2–10$\times$ speed-up at higher stiffness, while the Newmark–Beta method achieved comparable accuracy with a 6$\times$ reduction in runtime for the 1~MPa case.

In the helix experiments (Figure ~\ref{fig:comaprison_table}~(c,d)), we evaluated both low-stiffness and high-stiffness helices to further explore time-integration trade-offs. Each helix consisted of 100 nodes, $\rho = 1273.52~\text{kg/m}^3$, with $r = 1\times10^{-3}$~m for the 1~GPa case and $r = 5\times10^{-3}$~m for the 10~MPa case. In the low-stiffness regime, the explicit solver in PyElastica performed comparably to the Newmark–Beta scheme in Py-DiSMech, as the smaller time step size was offset by simpler per-step computations. However, Implicit Euler still outperformed it by nearly 7$\times$. In the high-stiffness regime, the Newmark–Beta scheme achieved a $>$10$\times$ speed-up, while Implicit Euler exceeded $>$30$\times$. These improvements were further enhanced by the use of the sparse solver PyPardiso, which reduced computation time substantially for large degrees-of-freedom due to edge twist angles in the 3D model. For accuracy comparison in the absence of analytical solutions, we applied damping and tracked the vertical displacement of the lowermost node in both simulators (Figures~\ref{fig:comaprison_table}~(e)–(f)), observing similar steady-state configurations.

In the snake locomotion experiment (Figure \ref{fig:comaprison_table}~(b)), we demonstrate that an implicit simulator can efficiently handle complex actuation dynamics. The snake comprised 101 nodes, $\rho = 1000~\text{kg/m}^3$, $r = 1.75\times10^{-3}$~m, and length $0.1$~m. As Py-DiSMech and PyElastica implement actuation via natural strains and internal muscle torques, respectively, we experimentally tuned Py-DiSMech’s strain profile to match PyElastica’s longitudinal actuation behavior. This achieved comparable forward locomotion with approximately a 2$\times$ speed-up. The actuation was applied as a traveling strain wave with a two-second ramping period, resulting in similar steady-state motion after the initial transient.

Overall, these experiments highlight that Py-DiSMech consistently achieves $2–30\times$ faster simulations than PyElastica while maintaining comparable accuracy. The combination of implicit integration, sparse solvers, and vectorized computation enables robust performance across stiffness regimes, demonstrating the scalability and efficiency of the proposed framework.

\section{Simulation Demonstrations}
\label{sec: results}
\begin{figure*}[t!]
\centerline{\includegraphics[clip, trim=0cm 10.5cm 0cm 0cm, width =\textwidth]{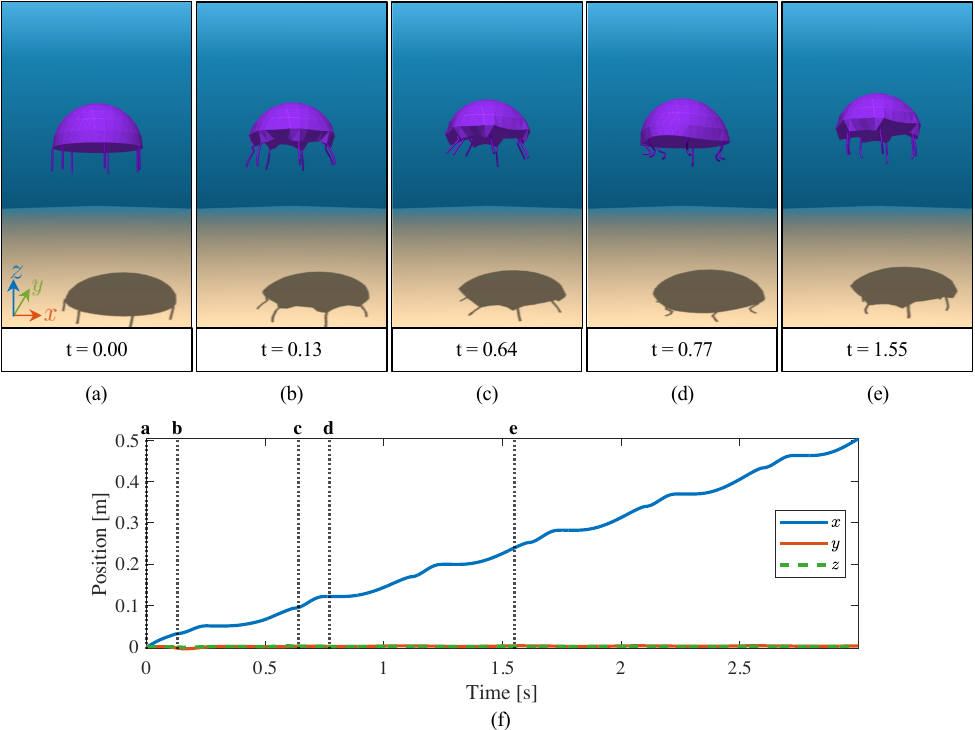}}
\caption{Jellyfish propelling upward through periodic undulation of the dome-shaped shell.(a-e) Snapshots of the Jellyfish at different time stamps. (f) Displacement of the topmost point of the jellyfish with time. The time stamps corresponding to the snapshots (a-e) are marked using vertical dotted lines.
}
\label{fig::jellyfish}
\end{figure*}

\begin{figure}[t!]
\centerline{\includegraphics[clip, trim=3cm 13.4cm 3cm 0cm, width=1\columnwidth]{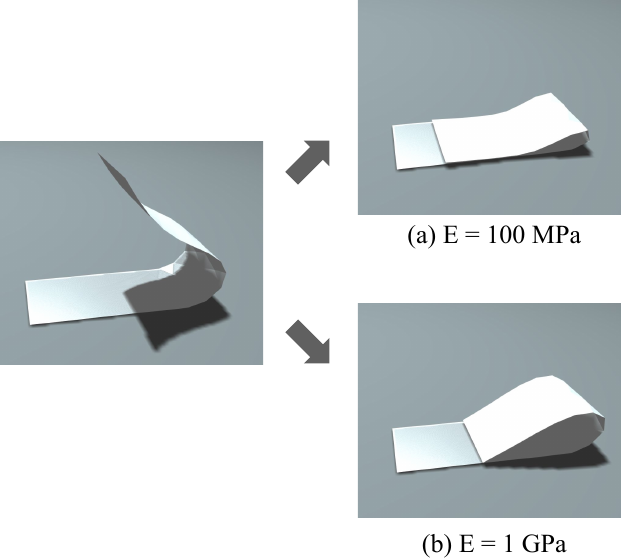}}
\caption{Rectangular shell folding onto itself under gravity. (a) Final configuration for a shell with Young’s modulus of 100 MPa. (b) Final configuration for a shell with Young’s modulus of 1 GPa. The stiffer shell exhibits a greater folded height compared to the more compliant one.
% \fix{}
}
\label{fig::shell_contact}
\end{figure}

% \begin{figure}[t!]
% \centerline{\includegraphics[clip, trim=2.7cm 12.7cm 2.7cm 0cm, width=1\columnwidth]{Figures/circle_folding_half_page.pdf}}
% \caption{Circular shell fixed at its center folding over itself under gravity. (a) Final configuration for a shell with Young’s modulus of 2~MPa. (b) Final configuration for a shell with Young’s modulus of 20~MPa. The softer shell exhibits a greater number of folds compared to the stiffer one.
% }
% \label{fig::circle_contact}
% \end{figure}

In this section, we present some showcase simulations using Py-DiSMech. In particular, we show the simulation of a jellyfish undulating and swimming upward, and two experiments showcasing self contact handling in shell: folding of a rectangular shell on itself and a circular shell held at the centre falling over itself and taking a shape with folds. All simulation examples are implemented as Jupyter notebooks, which are available in the repository referenced in the Code Availability section.

% \begin{figure*}[t!]
% \centerline{\includegraphics[clip, trim=0cm 10.5cm 0cm 0cm, width =\textwidth]{Figures/jellyfish.pdf}}
% \caption{Jellyfish propelling upward through periodic undulation of the dome-shaped shell.(a-e) Snapshots of the Jellyfish at different time stamps. (f) Displacement of the topmost point of the jellyfish with time. The time stamps corresponding to the snapshots (a-e) are marked using vertical dotted lines.
% }
% \label{fig::jellyfish}
% \end{figure*}

\textit{Jellyfish.} As our first demonstration, we simulate the undulating motion of a jellyfish bell, which results in upward propulsion. The jellyfish is modeled as a hemispherical shell with slender rods attached along its edge to represent the tentacles. The material density is set to 1100 kg/m$^3$, with Young’s moduli of 10 GPa for the bell and 10 MPa for the tentacles. A Poisson’s ratio of 0.5 is used for both components.
External forces considered in this simulation include hydrodynamic drag and thrust. The material density is selected such that gravitational and buoyant forces cancel each other out. Thrust is modeled as a force in the positive $z$-direction, proportional to the volume of water displaced due to the bell's contraction and expansion during undulation.
Figures~\ref{fig::jellyfish}(a–e) illustrate intermediate configurations of the jellyfish during the simulation. The displacement of the topmost point, plotted in Figure~\ref{fig::jellyfish}(f), reveals that the jellyfish undergoes periodic upward propulsion in the $z$ direction, while motion in the $x$ and $y$ directions remains negligible.
It is important to note that this simulation does not aim to replicate the precise biomechanics of a jellyfish; rather, it serves to demonstrate the simulator’s ability to reproduce biomimetic behaviors, which is of significant utility in the study and development of soft robotic systems.

% \begin{figure*}[t!]
% \centerline{\includegraphics[clip, trim=0cm 9cm 0cm 0cm, width =\textwidth]{Figures/jellyfish.pdf}}
% \caption{Jellyfish propelling upward through periodic undulation of the dome-shaped shell.(a-e) Snapshots of the Jellyfish at different time stamps. (f) Displacement of the topmost point of the jellyfish with time. The time stamps corresponding to the snapshots (a-e) are marked in the plot using vertical dashed lines.
% }
% \label{fig::jellyfish}
% \end{figure*}

\textit{Rectangular and circular plate folding.} 
Two examples are presented to demonstrate the simulator’s capability to robustly handle self-contact in thin shells using the proposed implicit contact formulation. The first case, shown in Figure~\ref{fig::shell_contact}, involves two rectangular shells with Young’s moduli of 100~MPa and 1~GPa, respectively, resting partially on the ground and folding onto themselves under gravity. As shown, both shells successfully fold, with the final folded height and overlap area dependent on the material stiffness. Consistent with physical intuition, the stiffer shell exhibits a greater folded height compared to the more compliant one.

\begin{figure}[t!]
\centerline{\includegraphics[clip, trim=2.7cm 12.9cm 2.7cm 0cm, width=1\columnwidth]{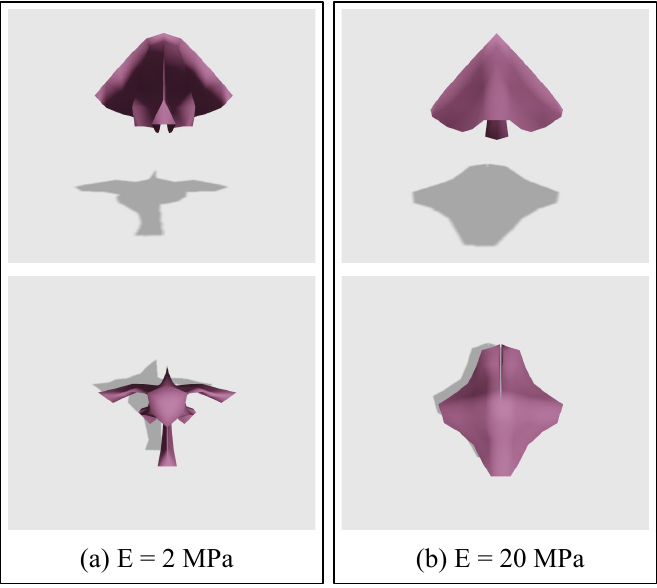}}
\caption{Circular shell fixed at its center folding over itself under gravity. (a) Final configuration for a shell with Young’s modulus of 2~MPa. (b) Final configuration for a shell with Young’s modulus of 20~MPa. The softer shell exhibits a greater number of folds compared to the stiffer one.
}
\label{fig::circle_contact}
\end{figure}

In the second case, a circular cloth modeled as a shell is fixed at its center and allowed to fall freely under gravity. As it descends, it naturally folds over itself. Figure~\ref{fig::circle_contact} illustrates two such examples with Young’s moduli of 2~MPa and 20~MPa. The softer cloth (2~MPa) undergoes more pronounced folding than the stiffer one, demonstrating the framework’s ability to capture complex self-contact dynamics across a wide range of material stiffnesses.

% \begin{figure}[t!]
% \centerline{\includegraphics[clip, trim=3cm 13.2cm 3cm 0cm, width=1\columnwidth]{Figures/paper_folding.pdf}}
% \caption{Rectangular shell folding onto itself under gravity. (a) Final configuration for a shell with Young’s modulus of 100 MPa. (b) Final configuration for a shell with Young’s modulus of 1 GPa. The stiffer shell exhibits a greater folded height compared to the more compliant one.}
% \label{fig::shell_contact}
% \end{figure}

% \begin{figure}[t!]
% \centerline{\includegraphics[clip, trim=2.7cm 12.7cm 2.7cm 0cm, width=1\columnwidth]{Figures/circle_folding_half_page.pdf}}
% \caption{Circular shell fixed at its center folding over itself under gravity. (a) Final configuration for a shell with Young’s modulus of 2~MPa. (b) Final configuration for a shell with Young’s modulus of 20~MPa. The softer shell exhibits a greater number of folds compared to the stiffer one.}
% \label{fig::circle_contact}
% \end{figure}

% \begin{figure*}[t!]
% \centerline{\includegraphics[clip, trim=0cm 9cm 0cm 0cm, width =\textwidth]{Figures/jellyfish.pdf}}
% \caption{Jellyfish propelling upward through periodic undulation of the dome-shaped shell.(a-e) Snapshots of the Jellyfish at different time stamps. (f) Displacement of the topmost point of the jellyfish with time. The time stamps corresponding to the snapshots (a-e) are marked in the plot using vertical dashed lines.
% }
% \label{fig::jellyfish}
% \end{figure*}

\section{Feedback Control}
\label{sec: control}

In this section, we demonstrate the efficacy of Py-DiSMech as a tool that aids in control design for soft robots. In particular, we design and implement feedback control laws to achieve a target shape (regulation problem) or a target trajectory (tracking problem) in dynamic soft-robotic structures. 
%
% \fix{One sentence reminder of the purpose of the software so that jump to feedback control does not seem abrupt.}
% This section describes the implementation of feedback control for achieving a target shape (regulation problem) or a target trajectory (tracking problem) in dynamic soft-robotic structures. 
%
Two representative case studies are presented: (1) a rod simply supported at both ends, where the objective is to attain a prescribed target shape while satisfying boundary constraints; and (2) a rod executing serpentine locomotion to follow a desired trajectory.

The control strategy adopted in the presented examples is based on the classical proportional–integral (PI) formulation applied to the natural strain fields of the structure. In the two-dimensional rod examples considered here, the control inputs are the natural longitudinal strain and curvature, denoted by $\bar{\boldsymbol{\epsilon}}$ and $\bar{\boldsymbol{\kappa}}$. The plant model is the discrete elastic rod (DER) formulation implemented in Py-DiSMech. The reference input corresponds to the desired strain field, which is computed from the reference state relative to the undeformed (original) configuration. The measured output is the actual state obtained from the simulation, from which the corresponding strain field is evaluated. The PI controller computes corrective increments based on the residual between the reference and measured strain fields. Spatial smoothing, rate limits on curvature change, and an anti-windup mechanism are incorporated to ensure stable operation.

\begin{figure}[t!]
\centerline{\includegraphics[clip, trim=0cm 7.95cm 0cm 0cm, width=1\columnwidth]{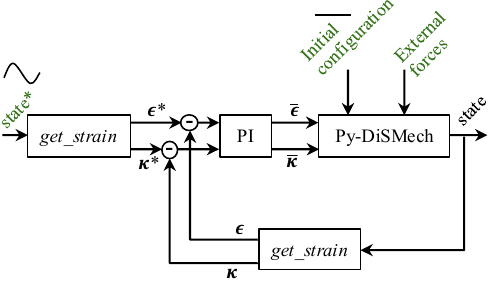}}
\caption{Feedback control architecture. The reference strain is computed from the desired configuration relative to the undeformed state, and the measured strain is obtained from the simulated configuration.}
\label{fig::control_blockdiagram}
\end{figure}

This natural-strain-based control provides a physically intuitive means of actuation, mirroring the behavior of soft actuators such as shape-memory alloys and PneuNet actuators, where changes in intrinsic strain drive deformation. The overall control architecture is illustrated in Figure~\ref{fig::control_blockdiagram}. Although a classical PI scheme is employed here for clarity and simplicity, the same framework is readily extensible to implement advanced model-based or optimal control methods—such as model predictive control or nonlinear feedback—using the same strain-space actuation interface.

\begin{figure*}[t!]
\centerline{\includegraphics[clip, trim=0cm 4.51cm 0cm 0cm, width =\textwidth]{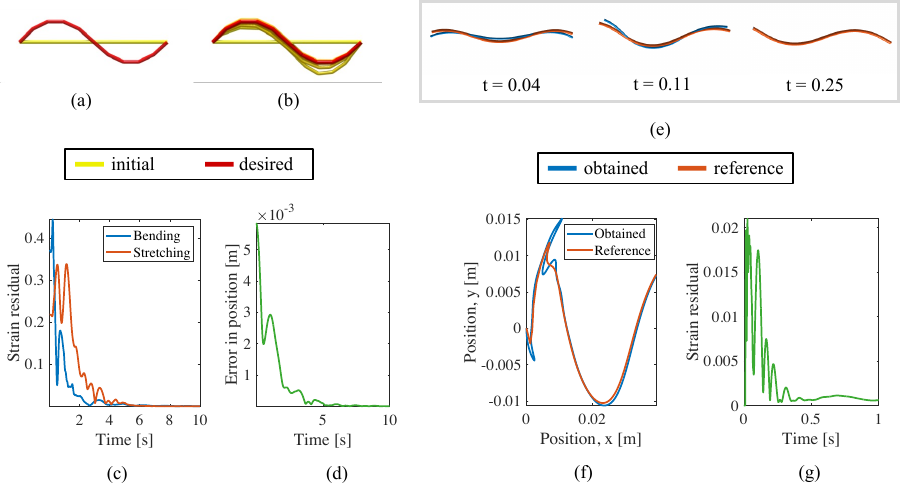}}
\caption{PI control for target shape regulation and trajectory tracking of a rod. (a-d) Target shape regulation: (a) Initial configuration and desired shape, (b) intermediate configurations, (c) evolution of strain residuals for bending and stretching modes, and (d) position error convergence over time. (e-g) Trajectory tracking for serpentine locomotion: (e) snapshots of intermediate configurations of the obtained rod configuration overlapped with the reference configurations, here time is given in seconds, (f) Obtained trajectory of the leading node of the rod overlapped with its reference trajectory, (g) plot of the strain residual with time. In both the cases--regulation and tracking--it can be seen that the desired configuration is achieved and the residuals decrease over time.
}
\label{fig::control}
\end{figure*}

\textit{Shape Regulation.}
For the first demonstration, a rod simply supported at both ends is actuated to achieve a prescribed horizontal S-shaped configuration, as shown in Figure~\ref{fig::control}~(a). The rod is subjected to gravity and viscous damping. Figures~\ref{fig::control}~(c) and~\ref{fig::control}~(d) show the time evolution of the strain residual and the shape error, quantified as the root-mean-square error (RMSE) between the nodal coordinates of the target and actual configurations, both converging to zero. Figure~\ref{fig::control}~(b) presents intermediate poses of the rod as it approaches the desired configuration. The results demonstrate accurate steady-state regulation and stability under the proposed feedback formulation.

% \begin{figure}[t!]
% \centerline{\includegraphics[clip, trim=0cm 5.2cm 0cm 0cm, width=1\columnwidth]{Figures/rod_s_control.pdf}}
% \caption{\fix{RADHA: Caption} Shape regulation of a simply supported rod.}
% \label{fig::rod_s_control}
% \end{figure}

\textit{Trajectory Tracking.}
In the second case, trajectory tracking is implemented to realize serpentine locomotion of a rod subjected to hydrodynamic drag forces computed using resistive-force theory. The control input is the natural curvature (bending strain), and the reference trajectory is obtained from a discrete elastic rod simulation with prescribed actuation. The reference curvature field is computed from the reference state and temporally interpolated to enable fine-grained tracking. Figure~\ref{fig::control}~(e) shows the intermediate configurations of the rod obtained through the feedback control against the reference configurations; the two are found to be almost overlapping about 0.25 sec onward. Figure~\ref{fig::control}~(h) compares the trajectory of the rod’s leading node with the reference path, illustrating accurate realization of the desired serpentine motion, while Figure~\ref{fig::control}~(i) shows the time evolution of the strain residual, indicating convergence to within 0.1\% of zero.

% \begin{figure}[t!]
% \centerline{\includegraphics[clip, trim=0cm 5.2cm 0cm 0cm, width=1\columnwidth]{Figures/rod_s_control.pdf}}
% \caption{Shape regulation of a simply supported rod using feedback control. (a) Initial configuration and desired shape, (b) intermediate configurations, (c) evolution of strain residuals for bending and stretching modes, and (d) position error convergence over time.}
% \label{fig::rod_s_control}
% \end{figure}

% \begin{figure}[t!]
% \centerline{\includegraphics[clip, trim=0cm 8.05cm 0cm 0cm, width=1\columnwidth]{Figures/snake_control.pdf}}
% \caption{\fix{RADHA: Caption, merge with Fig 12} Trajectory tracking in serpentine locomotion.}
% \label{fig::snake_control}
% \end{figure}

This straightforward yet effective formulation demonstrates the capability of Py-DiSMech for controller design, tuning, and validation in soft-robotic systems, highlighting its potential for both sim-to-real and real-to-sim integration.

\section{Conclusions}
\label{sec: conclusions}
This work presented Py-DiSMech, a Python-based, open-source simulation framework for modeling and control of soft robotic structures using the principles of Discrete Differential Geometry (DDG). By discretizing geometric quantities such as curvature and strain directly on meshes, Py-DiSMech enables accurate and efficient simulation of rods, shells, and their hybrid combinations. The framework integrates a fully vectorized numerical implementation for high performance, a penalty-energy-based implicit contact model capable of handling rod–rod, rod–shell, and shell–shell interactions, and a feedback control module for natural-strain-based actuation.

Quantitative comparisons demonstrated that Py-DiSMech achieves substantial speed-ups over existing geometry-based simulators while maintaining high physical fidelity. Showcase simulations—including jellyfish propulsion and folding of rectangular and circular shells—further highlight the framework’s ability to robustly capture large deformations and self-contact across a wide range of material stiffnesses. Finally, control examples demonstrated accurate target-shape regulation and trajectory tracking via classical PI feedback, underscoring Py-DiSMech’s suitability for controller design, tuning, and validation in dynamic soft-robotic systems.

Beyond its current implementation, Py-DiSMech’s modular and extensible design makes it well-suited for continued advancement. Future work will explore physics-informed differentiable energy learning for reduced-order elastic modeling and reinforcement learning for control optimization. Together, these directions will further enhance Py-DiSMech’s role as a versatile platform for simulation-driven design, optimization, and sim-to-real exploration in soft robotics.

\section*{Acknowledgments}

We acknowledge financial support from the National Science Foundation (award numbers: CMMI-2209782 and CAREER-2047663).

\section*{Code Availability}
All source code for the work presented in this paper is made available at {\url{https://github.com/StructuresComp/dismech-python}}.


\end{document}